\documentclass[11pt]{amsart}
\usepackage{geometry}                
\usepackage{graphicx}
\usepackage{amssymb}
\usepackage{epstopdf}
\usepackage[superscript,biblabel]{cite}
\title{A Misanthropic Reinterpretation of the Chinese Room Problem}
\author{Michael  Laufer}
\date{December 20, 2010}                                           

\begin{document}
\maketitle

The Chinese room problem, since being postulated by John Searle in 1980\cite{searle}, has been the basis of much discussion surrounding the ideas of consciousness, thought, understanding, and the mind-body problem. The argument asserts that if someone has a book which gives them rules for writing responses to queries in a language they do not understand, although they may have sufficient information to fool a native speaker into thinking they are corresponding with someone who understands, they do not understand the dialogue in the sense that they would if the same experiment were done with queries in the person's native language.

However, there has been very little exploration of the implied contrasting argument. When Searle states that \emph{he} understands English, and hence is able to understand the meaning of statements which are posed in English, and use that understanding to consciously construct responses, I am inclined to believe it, as he appears to me to be a creative individual, and a self-aware free thinker. On the other hand, I do not believe that this applies universally to all native speakers of English as he states.

I would suggest that many people do not understand, or think at all, during most of their waking life, but rather work via a large system of preprogrammed reactions, much the way he suggests that the artificial intelligences do not think.

How often have you witnessed someone decrying a political standpoint as wrong, or an entire genre of music as poor, merely because they had made the decision in advance? These instances are clearly moments when the understanding mechanism is not engaged.

To use a more mundane example, try to recall how often you have heard anyone reply to the ubiquitous query ``How are you?" after actually considering the question.

This of course raises the question of whether it is possible to distinguish between understanding, and the imitation of understanding. I put forward the following thought experiment: at a cocktail party there are two discussions going on, each one dominated by a single person. In one discussion, the dominating person is a scholar of some particular field. In the other discussion the discussion is dominated by someone who has no so-called ``understanding" of the topic, but has spent a sufficient amount of time around scholars of this particular field that they are able to imitate the behavior and speech of a scholar. So much so, that they have a sufficiently rich repertoire of responses to queries that any question we might pose is met with an answer which is thought provoking, and seemingly thoughtful. In a strictly scholarly context, we would label the latter person a plagiarist, rather than a scholar, but how would we ever be able to determine that? 

An immediate response is perhaps that we have faith that eventually the impostor would stumble, and state something which did not make sense, at which point they would be exposed. I ask, though, in the absence of such an event (either by richness of the impostor's repertoire, or by chance) how could we tell the difference between the two? Further, I ask, in the absence of such an error \emph{is} there a difference?

This question may have a zen \emph{tree-falls-in-the-forest} ring to it, however it strikes me as important; if an ignorant person says something deep, is it somehow less valuable? Certainly in our perception: we feel conned if we take a person to be a deep thinker based on a statement which is not theirs. The isolated statement, however is no less meaningful.

The most dangerous aspect of exploring this question, is the arising of the further question as to whether being a scholar is anything but a conditioned behavior pattern, and following trends of popular ideas.

The closest thing to an answer we might be a able to construct is that the scholar is capable of formulating new ideas, and the impostor is merely following a set of instructions. This brings us full circle, but our understanding is no greater. 

Just to be clear that I am not suggesting that I am above this: I feel many of my own actions are in the form of these preprogrammed reaction mechanisms. Indeed, they have to be; it would make very difficult work of going about one's daily business, if we each had to learn how to walk and speak anew again every morning. The preconditioned reactions are shortcuts our minds use in order to allow the brain to focus on other things. However, it seems that the mind is often lazy, and would much rather take a rule than explore an idea for soundness.

This may indeed explain why dogma is so appealing. A rule is a simple thing, following it lays no strain on the thought mechanism, and can be implemented straight away.

Even in more complicated matters, I find myself being an automaton: as a party trick, I can solve the Rubik's Cube, but as I do it, I am merely repeating a few algorithms, which allow me to move the cube from one state into another, but I do not understand what makes each of the algorithms perform its particular function. Perhaps it is merely that I haven't taken the time to examine them carefully enough to understand them, but even if I did, (or turned out to be incapable of doing so) it would not change my actions. 

Our sensibility labels the same actions as either genuine or fraudulent, based on whether there is an underlying understanding or not. The danger of this, I believe, is that if we begin exploring what fraction of human actions are genuine by this measure, we will find that fraction to be far lower than might be expected.

Think back to the schooling we all receive; it is predominantly a matter of amassing content, and procedural rulesets. Having a wealth of knowledge is often misconstrued as having an aptitude for thought, but really if all we are being taught to do is repeat thoughts of others, then schooling is training in \emph{not} thinking.

The Chinese room problem was constructed in hopes of showing how machines are not capable of understanding, while humans are. I think it sheds a frightening amount of light on the fact that whether we are or not, we certainly do very, very little of it.

The Sufis say that anyone who can go ten minutes without a conditioned response is illuminated. Frankly I think even ten seconds would be remarkable. Does the name Pavlov ring a bell?

\end{document}